\documentclass[sigconf]{acmart}
\AtBeginDocument{%
  }

\thanks{To appear in SAC 2026. This is the author's version of the work.}
\acmConference[SAC 2026]{The 41st ACM/SIGAPP Symposium On Applied Computing}{March 23-27, 2026}{Thessaloniki, Greece}

\usepackage{balance} 
\usepackage{microtype}
\usepackage{hyperref}
\usepackage{amsmath}
\usepackage{graphicx}
\usepackage{enumitem}
\usepackage{algorithm}
\usepackage{algorithmic}
\usepackage{booktabs}
\usepackage{verbatim}

\begin{document}

\title{Atomic Action Slicing: Planner-Aligned Options for Generalist VLA Agents}

\author{Stefan Tabakov}
\email{sitabakov@uni-sofia.bg}
\affiliation{%
  \institution  {Sofia University 'St. Kliment Ohridski'}
  \city{Sofia}
  \country{Bulgaria}
}

\author{Asen Popov}
\email{asepopov@tu-sofia.bg}
\affiliation{%
  \institution{Technical University of Sofia}
  \city{Sofia}
  \country{Bulgaria}
}

\author{Dimitar Dimitrov}
\email{d.dmitrov@student.utwente.nl}
\affiliation{%
  \institution{EEMCS, University of Twente}
  \city{Twente}
  \country{The Netherlands}
}

\author{S.~Ensiye Kiyamousavi}
\email{ensiye.kiyamousavi@gate-ai.eu}
\affiliation{%
  \institution{GATE Institute, Sofia University 'St. Kliment Ohridski'}
  \city{Sofia}
  \country{Bulgaria}
}

\author{Vladimir Hristov}
\email{vdhristov@tu-sofia.bg}
\affiliation{%
  \institution{Technical University of Sofia}
  \city{Sofia}
  \country{Bulgaria}
}

\author{Boris Kraychev}
\email{boris.kraychev@gate-ai.eu}
\affiliation{%
  \institution{GATE Institute, Sofia University 'St. Kliment Ohridski'}
  \city{Sofia}
  \country{Bulgaria}
}

\renewcommand{\shortauthors}{Tabakov et al.}

\begin{abstract}

  Current vision–language–action (VLA) models generalize poorly, particularly when tasks require new compositions of skills or objects. We introduce Atomic Action Slicing (AAS), a planner-aligned approach that decomposes long-horizon demonstrations into short, typed atomic actions that are easier for planners to use and policies to learn. Using LIBERO demonstrations, AAS produces a validated dataset of 2,124 atomic segments labeled with action type, temporal span, and confidence. A stronger segmenter (Gemini 2.5 Pro) closely matches planner-defined plans and remains robust under keyframe jitter, while smaller models perform worse on multi-object tasks. Fine-tuning CLIP-RT+ on our atomic dataset improves task success from 94.2\%→95.3\% on LIBERO-Goal and 83.8\%→88.8\% on LIBERO-Long. We publicly release the GATE-VLAP dataset on HuggingFace\footnote{https://huggingface.co/datasets/gate-institute/GATE-VLAP-datasets}.

\end{abstract}






\maketitle

\section{Introduction}

Generalist Vision--Language--Action (VLA) models promise unified perception, reasoning, and control across diverse tasks, environments, and embodiments.
Despite strong in-distribution results, recent systems - including \emph{OpenVLA}~\cite{kim2025openvla}, $\pi 0$~\cite{black2024pi0}, and follow-ups that add architectural or data scaling~\cite{zhao2025cotvla,li2024cogact,dey2025revla,qu2025spatialvla} - continue to degrade under out-of-distribution (OOD) tasks and novel compositions of skills.
A key driver is data bias: most demonstration corpora contain long-horizon behaviors with sparse semantic structure, making it difficult for monolithic policies to learn \emph{transferable} and \emph{composable} skills.

We propose \emph{Atomic Action Slicing (AAS)}, a pipeline that decomposes long-horizon videos (e.g., LIBERO) into short, planner-aligned atomic actions (temporally extended, verifiable skills), enabling direct integration of learned skills into classical planners.
Unlike prior skill-segmentation or option-discovery approaches, AAS bridges high-level planning with low-level learning by producing, to our knowledge, one of the first datasets of \emph{planner-aligned} robot demonstration segments, where each segment corresponds to a symbolic subtask.


We demonstrate two uses of this atomic skill set: (1) as operators for symbolic planners~\cite{fikes1971strips,erol1996complexity,nau2003shop2} (leaving a full planning complexity analysis to future work), and (2) for fine-tuning a VLA policy (CLIP-RT+~\cite{kang2024cliprt}) to improve compositional generalization.
Each atomic action in our work is treated as such an option, with verifiable success and symbolic postconditions to interface with a planner (cf.~\cite{konidaris2018skills}).

This paper makes the following contributions: 
  \textbf{Method:} AAS – a simple pipeline to split long demonstrations into short, typed atomic actions (three-stage process).
\textbf{Dataset:} A curated set of 2,124 planner-aligned atomic segments from LIBERO demos~\cite{liu2023libero}.
  \textbf{Improved Learning:} Fine-tuning CLIP-RT+~\cite{kang2024cliprt} on these segments boosts success rates (e.g., from 83.8\% to 88.8\% on LIBERO-Long).


\section{Related Work}

\textbf{Generalist VLA models:}
Recent Vision–Language–Action (VLA) policies combine large multimodal backbones with action heads to perform diverse manipulation tasks~\cite{kim2025openvla,black2024pi0,zhao2025cotvla,li2024cogact,dey2025revla,qu2025spatialvla}.
Although effective in-distribution, these monolithic models degrade on out-of-distribution (OOD) tasks and novel action compositions~\cite{spiridonov2025motovla,black2024pi0,kim2025openvla}, motivating approaches with explicit temporal structure instead of flat imitation.
\textbf{Learning from weakly labeled video:}
Large-scale human or web video has been used to learn visual and action priors without robot labels~\cite{nair2023r3m,radosavovic2023mvp,ma2023vip}.
Other work generates visual plans—future predictions or flow fields~\cite{du2023universal,xu2024flowinterface}—or learns discrete latent action codes for pretraining~\cite{ye2025lapa}.
These approaches provide motion priors but do not yield \emph{planner-grounded}, verifiable motor units.
\textbf{Hierarchical skills and options:}
Long-horizon tasks are often decomposed into reusable skills or motion primitives.
Geometry-based units such as point tracks and keypoints~\cite{bharadhwaj2024track2act,ren2025motiontracks,haldar2025pointpolicy} transfer across objects but lack typed preconditions and effects, limiting planner compatibility.
Options provide a natural planning interface when such symbolic structure is present~\cite{konidaris2018skills}.
\textbf{LLM-based planning and grounding:}
LLMs have been used to generate symbolic or subgoal-based task plans~\cite{ahn2022saycan,huang2022lmplanners,driess2023palme,karamcheti2024prismatic,zhen2024vla3d}, including systems such as VoxPoser~\cite{huang2023voxposer} and AutoGPT+P~\cite{birr2024autogptp}.
These methods supply high-level structure but still require reliable, executable low-level skills.

\emph{Positioning of our approach.} Our work operates in the context of benchmark tasks with formal schemas (e.g., BEHAVIOR Domain Definition Language - BDDL, LIBERO~\cite{srivastava2022behavior,liu2023libero}).
We extend these directions by extracting \emph{planner-aligned atomic actions} directly from demonstrations.
Our method anchors segmentation to symbolic plans, applies schema-constrained LLM predictions, and validates segments via count/order/duration checks.
The resulting dataset provides planner-ready operators and fine-grained supervision that improves compositional generalization beyond flat imitation~\cite{spiridonov2025motovla,kim2025openvla,black2024pi0}.

\section{Method}
\label{sec:method}

Long-horizon demonstrations often blur distinct skills, making it difficult for policies to learn reusable behaviors or for planners to rely on them. Our goal is to convert each trajectory into a sequence of \emph{atomic actions}: short, typed options with clear preconditions, effects, and temporal boundaries. These units form a symbolic operator set for planners and provide fine-grained supervision for learning.
\emph{Atomic Action Slicing} (AAS) produces these segments using three stages: (i) planner-guided \textbf{discovery}, (ii) \textbf{schema-constrained} LLM segmentation, and (iii) \textbf{validation} through count, order, and duration checks.
An example of an atomic action is given below.
\vspace{0.2cm}
\newcommand{\aaline}[2]{\makebox[3.2em][l]{\textbf{#1}}#2\par}

\aaline{Name}{\texttt{place\_bowl\_in\_drawer(bowl, drawer)}}
\aaline{Pre}{grasped(bowl), isOpen(drawer), clear(drawer)}
\aaline{Term.}{bowl released in drawer; end-effector exits}
\aaline{Post}{in(bowl, drawer), $\lnot$grasped(bowl)}
\aaline{Span}{30--120 frames; confidence $c\in[0,1]$}

\vspace{-0.2cm}

\vspace{0.5cm}
\textbf{Problem Formulation.} 
Consider an episode 
$\tau=(o_{1:T}, s_{1:T}, a_{1:T}, \ell, \mathcal{E})$. Here $o_t$ are images, $s_t$ robot or object states, and $a_t$ low-level actions. The instruction $\ell$ describes the task, while $\mathcal{E}$ gives a symbolic scene with objects, relations, and goals. The goal is to output validated atomic segments
\(
\hat{\Gamma}=\big[(\hat{o}_k, t^{(k)}_{\mathrm{start}}, t^{(k)}_{\mathrm{end}}, c^{(k)})\big]_{k=1}^{K},
\)
with labels $\hat{o}_k \in \Sigma$ taken from a typed schema (e.g., \texttt{open\_drawer}, \texttt{grasp}, \texttt{place}). Each label denotes an option $\langle \mathcal{I}_{\hat{o}}, \pi_{\hat{o}}, \beta_{\hat{o}}\rangle$. Symbolic preconditions $\mathrm{pre}(\hat{o})$ and effects $\mathrm{eff}(\hat{o})$ connect these options to STRIPS/HTN planners.


\vspace{0.3cm}
\textbf{Stage I: Discovery (Planner-Guided Decomposition).} \label{stage_I}
Given $(\ell,\mathcal{E})$, a task planner such as ~\cite{birr2024autogptp} returns an ordered list of atomic actions:
$P=(\hat{o}_1,\ldots,\hat{o}_K), \quad \hat{o}_k\in\Sigma.$
These subtasks fix the expected number of steps $K$ and their order. Semantic typing makes the plan consistent across episodes of the same task. Compared to change-point heuristics, planner guidance yields segments that are easier to reuse and verify.

\vspace{0.3cm}
\textbf{Stage II: Schema-Constrained LLM Segmentation.}\label{stage_II}
For each episode, a small set of keyframes $\mathcal{K}=\{t_i\}_{i=1}^{K_f}$ is selected, optionally with compact state summaries (e.g., gripper width or a joint angle). 
A multimodal vision language model receives four elements: the instruction and symbols from $\mathcal{E}$; the typed schema $\Sigma$ and ordered anchors $P$; a few-shot set; and simple temporal cues. The model proposes boundaries $\{(t^{(k)}_{s}, t^{(k)}_{e})\}_{k=1}^{K}$ under the constraints
{\small
\[
\begin{split}
t^{(1)}_{s}=1,\quad t^{(K)}_{e}=T,\quad t^{(k)}_{s}\le t^{(k)}_{e}, t^{(k)}_{e}+1=t^{(k+1)}_{s},
\quad \hat{o}_k\in\Sigma,\quad \hat{o}_k=P[k].
\end{split}
\]}
Contiguity, coverage, and label validity are therefore guaranteed, while the model focuses on boundary placement. 


\vspace{0.3cm}
\textbf{Stage III: Validation and Confidence Assignment.}\label{stage_III}
A candidate sequence $\tilde{\Gamma}$ is accepted only when three tests pass.
\begin{itemize}[leftmargin=1.2em]\itemsep0.2ex
  \item \textbf{Count:} the number of segments equals $K$.
  \item \textbf{Order:} the label sequence matches $P$, and times increase strictly.
  \item \textbf{Duration:} each span length $d_k=t^{(k)}_{e}-t^{(k)}_{s}+1$ lies in a class- and task-dependent range $[d_{\min}(\hat{o}_k), d_{\max}(\hat{o}_k)]$.
\end{itemize}
Accepted steps receive confidences $c^{(k)}\in[0,1]$. The score blends the model’s internal signal, slack to duration bounds, and agreement under keyframe jitter. High-precision subsets arise by thresholding on $c^{(k)}$.
\vspace{0.3cm}
\textbf{Planner--Learner Interfaces.}
Two downstream interfaces follow. \textbf{Planning.} STRIPS/HTN can operate over the option alphabet; measuring search complexity and plan repair with these operators is left to future work. \textbf{Learning:} labeled windows feed policy or representation learning. This enables hierarchical training and compositional evaluation.








\begin{algorithm}[h]
\scriptsize
\caption{Atomic Action Slicing (Planner-Aligned)}
The key inputs are the task spec (goal and environment), demonstration data, and the LLM/VLM model for segmentation. The outputs are the segments.
\label{alg:aas}
\begin{algorithmic}[1]

\REQUIRE Episodes $\{\tau_i=(o^{(i)}_{1:T_i}, s^{(i)}_{1:T_i}, \ell^{(i)}, \mathcal{E}^{(i)})\}$, schema $\Sigma$, few-shot set $\mathcal{F}$, keyframe budget $K_f$
\FOR{each episode $\tau$}
  \STATE \textbf{Discovery:} Obtain atomic plan $P=(\hat{o}_1,\ldots,\hat{o}_K)$ from planner given $(\ell,\mathcal{E},\Sigma)$
  \STATE \textbf{Segmentation (schema-constrained):} 
  Select keyframes $\mathcal{K}=\{t_1,\ldots,t_{K_f}\}$ and extract optional state summaries
  \STATE Query VLM with $\{\ell,\mathcal{E},\Sigma,P,\mathcal{F},\mathcal{K}\}$ to propose
         $\{(t^{(k)}_{s}, t^{(k)}_{e})\}_{k=1}^{K}$ subject to contiguity and coverage
  \STATE \textbf{Validation:} If \textsc{Count}($K$), \textsc{Order}($P$), and \textsc{Duration}($(t^{(k)}_{s}, t^{(k)}_{e})$) all hold:
  \STATE \hspace{1.5em} set $c^{(k)} \leftarrow \textsc{Calibrate}(\text{VLM\_score}^{(k)},$
\STATE \hspace{5em} $\text{duration\_slack}^{(k)}, \text{jitter\_agreement}^{(k)})$
  \STATE \hspace{1.5em} append $(\hat{o}_k, t^{(k)}_{s}, t^{(k)}_{e}, c^{(k)})$ to $\hat{\Gamma}$
  \STATE \textbf{If not valid:} optionally refine prompt/keyframes and re-query specific steps
\ENDFOR
\STATE \textbf{return} validated segments $\{\hat{\Gamma}\}$ for planning and learning
\end{algorithmic}
\end{algorithm}
\vspace{-0.3cm}

\section{Evaluation}
\label{sec:eval}

We evaluate Atomic Action Slicing (AAS) on LIBERO using two experiments: \ref{sec:eval:palign} plan alignment and stability, and \ref{sec:eval:finetune} fine-tuning with atomic segments. All runs use AutoGPT+P for discovery and the schema-constrained segmenter for alignment (Alg.~\ref{alg:aas}). 

\subsection{Common Setup and Metrics}
\emph{Data.}
We select LIBERO manipulation tasks 
(e.g., \texttt{libero\_10 
/ put\_bowl\_in\_drawer}) 
and subsample $N$ demonstrations. Each episode provides video frames and synchronized JSON logs (robot states/actions). BDDL descriptors supply object types and goal symbols.

\noindent
\emph{Planner and schema.}
Discovery queries AutoGPT+P with the BDDL scene and instruction to obtain an ordered atomic plan $P=(\hat{o}_1,\dots,\hat{o}_K)$ from a typed schema $\Sigma$ (\texttt{open\_drawer}, \texttt{grasp}, \texttt{place}, \dots).

\noindent
\emph{Predicate-free sequence and temporal metrics.}
We deliberately avoid task-specific predicates in Exp.~\ref{sec:eval:palign}. Let $\hat{y}_{1:K}$ be the predicted label sequence and $y_{1:K}$ the planner sequence.


\begin{enumerate}[leftmargin=1.2em,itemsep=0pt,topsep=2pt]
\item \textbf{SeqAcc}: $\mathbb{I}[\hat{y}_{1:K}{=}y_{1:K}]$
\item \textbf{EditSim}: $1-\mathrm{Lev}(\hat{y},y)/\max(K,|\hat{y}|)$
\item \textbf{Cnt/Ord}: $|\hat{y}|{=}K$ and ($\hat{y}$ matches the planned order)
\item \textbf{IoU}\textsubscript{idx}: $\tfrac{1}{K}\sum_k\mathrm{IoU}([\hat{t}^k_s,\hat{t}^k_e],[t^k_s,t^k_e])$
\item \textbf{MAE}\textsubscript{start,end,dur}: $\tfrac{1}{K}\sum_k|\cdot|$
\item \textbf{Stability@Jitter}: $\frac{1}{K}\sum_{k}\mathrm{IoU}\big([t^{(0)}_{s,k},t^{(0)}_{e,k}], [t^{(2)}_{s,k},t^{(2)}_{e,k}]\big)$
\end{enumerate}
where $\mathrm{IoU}$ is interval IoU on frames and superscripts $(0)$ vs.\ $(2)$ denote two runs of the same model with keyframe jitter 0 vs.\ $\pm2$ frames.

\begin{table}[h]\small
\vspace{-0.5cm}
\centering
\begin{tabular}{lccc}
\toprule
\textbf{Metric} & \textbf{Flash} & \textbf{Pro} & $\Delta$ (Pro $-$ Flash) \\
\midrule
Successful segmentations (out of 100) & 74 & \textbf{93} & +19 \\
Success rate & 74.0\% & \textbf{93.0\%} & +19 pp \\
Avg. segments per demo & 3.41 & 3.46 & +0.05 \\
Avg. trajectory length (frames) & 268.1 & 267.0 & $-$1.1 \\
Mean Kendall’s $W$ & 0.9105 & \textbf{0.9136} & +0.0031 \\
\bottomrule
\end{tabular}
\caption{\small Overall segmentation performance of Gemini~2.5~Flash vs. Gemini~2.5~Pro on 100 LIBERO demonstrations.}
\vspace{-1.5cm}
\end{table}

\subsection{Experiment 1: Plan Alignment and Segmentation Consistency}
\label{sec:eval:palign}
\paragraph{Goal:} Evaluate how well a vision-language model can segment long-horizon demonstrations into correct atomic actions, and compare a lightweight model (Gemini~2.5~Flash) to a stronger model (Gemini~2.5~Pro) on consistency and alignment with expected subtasks. This experiment tests if zero-shot segmentation by large models can reliably reproduce the intended sequence of actions in a task.

\emph{Results:} Both models successfully segmented the majority of the 100 LIBERO demonstrations, with Gemini~2.5~Pro achieving a 93\% segmentation success rate (93/100 demos) versus 74\% for Flash. On average, both produced about 3.4 segments per demo, closely matching the expected number of subtasks and indicating neither under-segmentation nor over-segmentation. Table~1 summarizes the overall performance. Notably, Gemini~2.5~Pro showed near-perfect agreement with the planner-defined subtask sequence (Sequence Accuracy $\approx 1.0$ and Edit Similarity $\approx 1.0$), while Flash was slightly less aligned on complex multi-object tasks. Both models exhibited high consistency: the inter-model concordance measured by Kendall’s $W$ was above $0.91$ for each, comparable to human annotator agreement on similar tasks. This suggests that the segmentation boundaries chosen by the models are largely repeatable and reliable.

\subsection{Experiment 2: Fine-Tuning VLA Agents with Atomic Segments}
\label{sec:eval:finetune}
\paragraph{Goal:} We investigate whether training on planner-aligned atomic actions can improve a policy’s generalization on downstream tasks. In this experiment, we fine-tune a state-of-the-art VLA model (the CLIP-RT+ robot policy from \cite{kang2024cliprt}) using our segmented subtasks – the Atomic Action Dataset – and evaluate its performance on the LIBERO benchmark. The Atomic Action Dataset is constructed by decomposing original demonstration trajectories into verifiable subtasks. For example, the long-horizon LIBERO tasks (e.g. “put the bowl in the drawer and close it”) are split into meaningful units like pick up bowl, place bowl in drawer, close drawer. This segmentation dramatically increases the number of training instances: LIBERO-Goal tasks (originally 434 demos) yield 758 atomic segments, and LIBERO-Long tasks (391 demos) produce 1366 segments, for a total of 2124 segmented subtasks. 


\emph{Results}: Fine-tuning on atomic segments yields a measurable boost in performance. The CLIP-RT+AA agent (with atomic-action supervision) achieves the highest overall success rate on LIBERO, outperforming the strongest prior baseline by roughly 2 percentage points on average. Notably, on the LIBERO-Goal tasks (shorter, goal-conditioned scenarios), the success rate increased from 94.2\% to 95.3\%, and on the challenging LIBERO-Long tasks (multi-step horizons) it rose from 83.8\% to 88.8\%. Table~\ref{tab:libero_success} breaks down the CLIP-RT+AA success rates before vs. after fine-tuning. We report the original CLIP-RT+ baseline performance alongside the improved results obtained with our atomic training.

\begin{table}[h]
\small
\centering
\begin{tabular}{lcc}
\toprule
\textbf{Task Suite} & \textbf{Original Demos} & \textbf{Atomic Segments} \\
\midrule
LIBERO-Goal& 434 & 758 \\
LIBERO-Long & 391 & 1366 \\
\textit{Combined Total}             & 825 & 2124 \\
\bottomrule
\end{tabular}
\caption{\small Dataset size before vs. after planner-aligned segmentation.}
\label{tab:dataset_size}
\vspace{-0.8cm}
\end{table}

\begin{table}[h]
\small
\centering
\begin{tabular}{lcc}
\toprule
\textbf{Task Suite} & \textbf{Baseline CLIP-RT+}& \textbf{Fine-tuned CLIP-RT+AA} \\
\midrule
LIBERO-Goal& 94.2\% & 95.3\% (+1.1\%)\\
LIBERO-Long& 83.8\% & 88.8\% (+5.0\%)\\
\bottomrule
\end{tabular}
\caption{\small CLIP-RT+AA success rates on LIBERO tasks before vs. after atomic fine-tuning.}
\label{tab:libero_success}
\vspace{-1cm}
\end{table}


\section{Discussion and Limitations}
\label{sec:discussion}

Our experiments demonstrate that Atomic Action Slicing (AAS) can reliably decompose long-horizon LIBERO demonstrations into short, planner-aligned atomic actions. The resulting segments exhibit (i) tight alignment with symbolic task plans, yielding accurate and stable temporal boundaries, (ii) semantic consistency through a typed action schema that keeps label sequences comparable across episodes of the same task, and (iii) a validated atomic dataset of 2{,}124 segments (758 from LIBERO-Goal and 1{,}366 from LIBERO-Long) with labels, time spans, and confidence scores. When used for downstream learning, this atomic supervision improves policy performance: fine-tuning CLIP-RT+ on our segments leads to an absolute increase in success rate (e.g., from 94.2\% to 95.3\% on LIBERO-Goal), indicating that planner-aligned options help agents acquire more robust, composable skills.

Despite these benefits, AAS has several limitations. First, it currently depends on structured environment descriptions in BDDL to generate the task plan, which restricts applicability in settings without rich symbolic specifications or where the scene description is incomplete. Second, the quality of temporal alignment remains sensitive to keyframe selection and video quality: if important transitions occur between sampled frames, or in very noisy sequences, the inferred boundaries can drift. Third, our evaluation is confined to LIBERO simulation; we have not yet validated the pipeline on real robot data or in more open-world environments~\cite{radosavovic2023mvp}.

\section{Conclusion}

We presented \emph{Atomic Action Slicing} (AAS), a simple planner-aligned method that splits long robot demonstrations into small, typed \emph{atomic actions}. These actions are easy for planners to use and useful for training policies. From LIBERO, we built a training-ready dataset of \textbf{2{,}124} atomic segments. This gives a practical bridge between symbolic plans and low-level control. 
Our experiments show two key results. (1) A strong segmenter matches expert plans with near-perfect sequence metrics and stays stable under small input changes. (2) Fine-tuning \textbf{CLIP-RT+} on our atomic dataset improves task success: \textbf{95.3\%} on LIBERO-\emph{Goal} (from 94.2\%) and \textbf{88.8\%} on LIBERO-\emph{Long} (from 83.8\%). Overall, AAS delivers planner-ready skills, a clean dataset for learning, and measurable gains in task completion and generalization for VLA agents.  

\begin{acks}
This research is part of the GATE project funded by the Horizon 2020 WIDESPREAD-2018-2020 TEAMING Phase 2 programme under grant agreement no. 857155, the programme "Research, Innovation and Digitalization for Smart Transformation" 2021-2027 (PRIDST) under grant agreement no. BG16RFPR002-1.014-0010-C01, and the European Regional Development Fund within the Operational Program “Bulgarian national recovery and resilience plan” and the procedure for the direct provision of grants “Establishing of a network of research higher education institutions in Bulgaria” under Project BG-RRP-2.004-0005 “Improving the research capacity and quality to achieve international recognition and resilience of TU-Sofia (IDEAS)”.

\end{acks}



\bibliographystyle{ACM-Reference-Format} 
\bibliography{atomic_actions_vla_refs}

\end{document}